\documentclass{article} 
\usepackage{iclr2027_conference,times}

\usepackage{booktabs}
\usepackage{tabularx}
\usepackage{array}
\usepackage{placeins}

\usepackage{xcolor}
\usepackage{graphicx}   

\usepackage{amsmath,amsfonts,bm}

\def\eqref#1{equation~\ref{#1}}

\def\1{\bm{1}}

\DeclareMathAlphabet{\mathsfit}{\encodingdefault}{\sfdefault}{m}{sl}
\SetMathAlphabet{\mathsfit}{bold}{\encodingdefault}{\sfdefault}{bx}{n}

\usepackage{hyperref}
\usepackage{url}
\iclrfinalcopy

\title{Sparse Autoencoders Can Induce Local Geometry for Mitigating Catastrophic Forgetting}

\author{Kamil Faber\thanks{Equal contributions} \\
Faculty of Computer Science\\
AGH University of Krakow\\
Krakow, Poland\\
\texttt{kfaber@agh.edu.pl} \\
\And
Katarzyna Filus\textsuperscript{*} \& {\L}ukasz Pawela \\
Institute of Theoretical and Applied Informatics, \\
Polish Academy of Sciences \\
Gliwice, Poland \\
\texttt{\{kfilus,lpawela\}@iitis.pl} 
}

\begin{document}

\maketitle
\lhead{Preprint}
\begin{abstract}

Catastrophic forgetting remains a challenge in supervised continual learning, as adapting to new tasks causes representation drift and loss of previously learned information. This raises the question of which representational changes should be constrained while preserving plasticity. We therefore introduce  \emph{Sparse Autoencoder Geometric Anchoring (SAGA)}, which converts a Sparse Autoencoder (SAE) trained at each task boundary into local, low-rank constraints on representation drift. Replay examples specify where the constraints apply, while leading right singular directions of the encoder Jacobian specify which changes are penalized. The resulting geometry is input dependent and can be retained without keeping the SAE for subsequent tasks' training. We establish its local relation to latent code matching, define its spectral truncation error, and derive a conditional bound on frozen-readout drift. With replay and snapshot history, the method achieves the highest mean Class-IL accuracy and lowest mean forgetting among the evaluated ER, DER++, ER-ACE, and SGD baselines on three pretrained ViT-B/16 benchmarks, with accuracy gains of 12.15, 4.79, and 3.22 percentage points over the strongest conventional baseline on CIFAR-100, CUB-200, and ImageNet-R, respectively. From scratch results are more mixed, while rank ablations and geometry controls show that SAE-induced low-rank geometry offers useful stability-plasticity trade-offs.

\end{abstract}

\section{Introduction}

Catastrophic forgetting, i.e., the degradation of previously acquired knowledge
during sequential training on new tasks, remains one of the key challenges in
supervised continual learning~\cite{wang2024comprehensive}. Common mitigation
strategies include rehearsal~\cite{rolnick2019experience}, regularization and
distillation~\cite{kirkpatrick2017overcoming,li2017lwf}, and hybrid
approaches~\cite{buzzega2020der}. More recently, forgetting has increasingly
been studied at the representation level, where learning subsequent tasks can
alter feature-space geometry and move previously useful information into
directions that are no longer accessible to the original
readout~\cite{gu2023preserving,filus2026lost,masip2026putting}. This motivates
preserving task-relevant representation subspaces, for example those derived
from previous task heads~\cite{yamashita2026rowspace} or
representations~\cite{saha2021gradient}. However, these approaches typically
define global subspaces, leaving open whether local geometry could provide 
selective control of representation drift.

Sparse autoencoders (SAEs) map dense representations to sparse feature
codes~\cite{huben2023sparse,bereska2024mechanistic,filus2026measuring}. In
continual learning, they have been used to analyze changes in visual
representations~\cite{filus2026lost,masip2026putting} and to regularize sparse
activations during language-model
adaptation~\cite{zhang2026sae,ning2026weights}. We ask a different
question: \emph{can a sparse encoder specify which local changes in the original
representation should be penalized, rather than only providing a latent code to
match?}

To address this question, we introduce \emph{SAGA} (\emph{Sparse Autoencoder Geometric Anchoring}), which uses SAE encoder Jacobians to derive task-specific local geometry for representation anchoring. At each task boundary,
a fresh SAE is fitted to the backbone representations of a  replay buffer
snapshot. Its encoder Jacobian defines a local quadratic geometry at each stored
example. The snapshot specifies \emph{where} to anchor the old representation, while
the leading Jacobian singular directions and their weights specify \emph{which
changes} to penalize. Unlike a single global subspace, these constraints can
differ across examples because the SAE's active features change with the input.
Once these elements are computed, the SAE can be discarded.
Full-stack SAGA combines this Jacobian anchor with two components.
\emph{History} keeps the anchors from all task boundaries, so older constraints
are not overwritten by snapshots fitted to already-drifted representations.
\emph{Replay} adds cross-entropy on snapshot examples, supplying the classifier
supervision that anchoring alone lacks in Class-IL.

The empirical benefit is clearest with pretrained representations. On
CIFAR-100, CUB-200, and ImageNet-R, full-stack \emph{SAGA} configuration achieves the highest mean accuracy and lowest mean
forgetting among the evaluated conventional baselines. Against the
corresponding latent-anchoring configuration, it has higher mean Class-IL
accuracy in four out of five dataset/backbone settings, with CIFAR-100 being the
exception. Rank ablations show that a small retained subspace can be sufficient, and
geometry controls identify where SAE-derived constraints help relative to global
alternatives. We distinguish these empirical findings from a universal
optimality claim: the mathematical analysis states the coverage of old-readout
sensitivity needed for a geometric anchor to control functional drift. Our contributions can be summarized in the following way:

\begin{itemize}
    \item A task-boundary anchoring method that turns sparse encoders into
    input-dependent, low-rank quadratic constraints, separating spatial memory
    in replay examples from directional memory in Jacobian factors.
    \item An analysis of the relation to latent matching, spectral truncation,
    and frozen-readout drift, with an explicit distinction between geometric
    energy and the normalization of the implemented training losses.
    \item An empirical evaluation across pretrained and from scratch backbones,
    including replay/history ablations, rank compression, alternative
    geometries, and dictionary size, that characterizes both the benefits and
    the limits of SAE-derived anchoring.
\end{itemize}

\section{Related work}

Traditional research on catastrophic forgetting in supervised continual learning has primarily focused on classifier-level degradation and mitigation strategies such as replay, regularization, distillation, and architectural approaches~\cite{kirkpatrick2017overcoming,li2017lwf,parisi2019continual,wang2024comprehensive,verwimp2024continual}. More recently, attention has shifted toward representations, including analyses that distinguish information loss from loss of accessibility to an old readout~\cite{davari2022probing}, and methods that explicitly preserve representation structure. LUCIR constrains feature directions~\cite{hou2019lucir}, PODNet distills spatially pooled representations~\cite{douillard2020podnet}, and Backward Feature Projection preserves linear separability of past tasks~\cite{gu2023preserving}. Related works preserve task-relevant subspaces defined from activations~\cite{saha2021gradient} or classifier weights~\cite{yamashita2026rowspace}. At the parameter level, OGD constrains updates using gradients of old outputs~\cite{farajtabar2020orthogonal}, while compressed quadratic constraints have been considered through sketched structural and Jacobian regularization~\cite{li2021sketched,heckel2022provable}. Thus, neither subspace nor Jacobian-based regularization is new in isolation. In contrast to these global or parameter-space constraints, our method uses the Jacobian of a task-anchored SAE encoder with respect to the backbone representation to induce an input-dependent, low-rank geometry that adapts to each replay anchor.

The use of encoder Jacobians to capture local representation geometry has also been studied in earlier work on autoencoders. In particular, the Manifold Tangent Classifier extracts local directions from top singular vectors of a contractive autoencoder encoder Jacobian and encourages classifier invariance along them~\cite{rifai2011manifold}. Our objective is different: rather than interpreting these directions as manifold tangents and enforcing classifier invariance, we freeze SAE sensitivity factors at task boundaries and use them to constrain subsequent backbone representation drift in continual learning. We make no assumption that the selected TopK-Jacobian directions correspond to manifold tangents or individual semantic concepts. Our sparse encoders build on $k$-sparse autoencoders~\cite{makhzani2013ksparse} and subsequent TopK SAE formulations~\cite{gao2025scaling}. Jacobian Sparse Autoencoders~\cite{farnik2025jacobian} address a complementary problem of learning sparse computational connections between model representations, rather than deriving continual-learning constraints from the local geometry of a frozen sparse encoder.

Recently, Sparse Autoencoders (SAEs), which map dense representations into sparse feature vectors~\cite{huben2023sparse,bereska2024mechanistic,filus2026measuring}, have been applied to continual learning both to analyze and mitigate forgetting. Diagnostic studies~\cite{filus2026lost,masip2026putting} show that forgetting can reflect changes in feature accessibility and geometry rather than complete information loss, with a substantial component of representational drift being approximately linear. For mitigation in Low-Rank Adaptation (LoRA)-based continual learning for LLMs, SAE-FD~\cite{zhang2026sae} uses directional feature matching and magnitude preservation, while SAE-guided activation regularization~\cite{ning2026weights} selectively protects SAE features. In contrast, we study supervised continual learning in vision and train \emph{task-boundary SAEs} rather than relying on a shared pretrained SAE space. Instead of directly preserving SAE activations, we export their local sensitivity geometry as low-rank factors and constrain representation drift along its dominant directions, without retaining the SAE during subsequent training. Our \emph{Latent} baseline therefore serves as a direct TopK-code matching ablation rather than a reproduction of these prior methods.


While subspace regularization, Jacobian-based geometry, and SAE-based preservation have each been explored in different contexts, our approach introduces \emph{task-boundary, input-dependent SAE geometry} for continual representation anchoring. At each task boundary, the SAE is converted into local, low-rank sensitivity directions that constrain subsequent representation drift and can be retained without keeping the SAE itself. Comparisons with latent anchoring and random, PCA, and classifier SVD geometries isolate the role of the resulting SAE-induced geometry.

\section{Using SAE-induced local geometry for forgetting mitigation}

\paragraph{Spatial and directional anchors.}
Let $s$ index a completed task and let $B_s=\{x_{s,i}\}_{i=1}^{n_s}$ be the
reservoir-buffer snapshot stored at that boundary. It contains examples from the
cumulative stream, not exclusively from task $s$. With backbone parameters
$\theta_s$, store $h_s(x)=f_{\theta_s}(x)\in\mathbb R^d$ and fit a fresh TopK
SAE encoder $E_s$~\cite{makhzani2013ksparse,gao2025scaling} to the snapshot representations. At each stored example,
compute
\begin{equation}
    J_s(x)=J_{E_s}(h_s(x))=U_s(x)\Sigma_s(x)V_s(x)^\top,
    \qquad \delta_s(x;\theta)=f_\theta(x)-h_s(x).
\end{equation}
Define the full pullback semimetric and its spectral truncation by
\begin{align}
\label{eq:full_Qs}
    Q_s(h)&=J_{E_s}(h)^\top J_{E_s}(h),\\
\label{eq:rank_Qs}
    Q_{s,r}(x)&=V_{s,r}(x)\Sigma_{s,r}(x)^2V_{s,r}(x)^\top\succeq0,
\end{align}
where $V_{s,r}$ contains the $r$ retained right singular vectors. The per-example
\emph{geometric energy} and its snapshot average are
\begin{align}
\label{eq:jacobian_geometric_energy}
    G_{s,r}(x;\theta)&=\|\Sigma_{s,r}(x)V_{s,r}(x)^\top\delta_s(x;\theta)\|_2^2
      =\delta_s(x;\theta)^\top Q_{s,r}(x)\delta_s(x;\theta),\\
    \widehat R_{s,r}(\theta)&=\frac{1}{|B_s|}\sum_{x\in B_s}G_{s,r}(x;\theta).
\end{align}
The snapshot specifies \emph{where} the old representation is anchored and
$Q_{s,r}(x)$ specifies \emph{which local changes} are penalized. In Jacobian
mode, the encoder is discarded after these factors have been computed; the
cached representations and factors define the subsequent constraint.
Appendix~\ref{appsec:sae_setup} specifies the architecture and construction.

\paragraph{Training losses and snapshot history.}
For an anchor minibatch $\mathcal M_s\subset B_s$, the raw Jacobian loss
averages over examples \emph{and} retained directions, whereas the latent
baseline sums latent-coordinate errors and averages over examples:
\begin{align}
\label{eq:implemented_jacobian_loss}
    \mathcal L_{\mathrm{Jac},s}^{\mathrm{impl}}
      &=\frac{1}{|\mathcal M_s|r}\sum_{x\in\mathcal M_s}G_{s,r}(x;\theta),\\
\label{eq:implemented_latent_loss}
    \mathcal L_{\mathrm{latent},s}^{\mathrm{impl}}
      &=\frac{1}{|\mathcal M_s|}\sum_{x\in\mathcal M_s}
        \|E_s(f_\theta(x))-E_s(h_s(x))\|_2^2.
\end{align}
Thus the raw training metric is $Q_{s,r}/r$, not $Q_{s,r}$. \texttt{Vanilla}
retains the latest snapshot; \texttt{History} retains all completed-task
snapshots. Let $\mathcal H_t$ be this retained index set when learning task $t$.
An independent minibatch is drawn from each snapshot and the default mean
reduction is
\begin{equation}
\label{eq:history_impl}
    \mathcal L_{\mathrm{anchor}}^{\mathrm{impl}}
      =\frac{1}{|\mathcal H_t|}\sum_{s\in\mathcal H_t}\mathcal L_s^{\mathrm{impl}}.
\end{equation}
\texttt{Replay} adds cross-entropy on the minibatch from the \emph{latest frozen
snapshot}, rather than drawing from the live reservoir. With $\lambda$ the
anchor coefficient and $\rho$ the replay coefficient, the objective is
\begin{equation}
\label{eq:training_objective}
    \mathcal L=\mathcal L_{\mathrm{stream\ CE}}
      +\lambda\mathcal L_{\mathrm{anchor}}^{\mathrm{impl}}
      +\rho\mathcal L_{\mathrm{snapshot\ CE}}.
\end{equation}
Without \texttt{Replay}, $\rho=0$. Before the first snapshot is created, only
the stream loss is used. Stored images have deterministic preprocessing for
anchor comparisons so that the old and current features refer to the same input
view.

\paragraph{Relation to latent matching and rank truncation.}
For $E(h)=\operatorname{TopK}(\operatorname{ReLU}(Wh+b))$, TopK retains the $K$
largest post-ReLU values in their original coordinates. If the positive active
mask is unchanged along $h_0+\tau\delta$, $0\leq\tau\leq1$, then
\begin{equation}
    E(h_0+\delta)-E(h_0)=J_E(h_0)\delta.
\end{equation}
At full retained rank, the unnormalized Jacobian energy therefore equals latent
displacement energy. The \emph{implemented} raw losses satisfy $\mathcal
L_{\mathrm{latent}}^{\mathrm{impl}} =r\mathcal L_{\mathrm{Jac}}^{\mathrm{impl}}$
within a fixed cell, so equal coefficients do not give equal local
regularization strength. Our latent comparison concerns these implemented
objectives, not a matched-strength isolation of support geometry.

For $Q=J^\top J$ and $Q_r=\sum_{i=1}^r\sigma_i^2v_iv_i^\top$, spectral
truncation is optimal in operator and Frobenius norms and gives
\begin{equation}
    0\leq\delta^\top(Q-Q_r)\delta
      \leq\sigma_{r+1}^2\|\delta\|_2^2,
    \qquad \dim\ker Q_r\geq d-r.
\end{equation}
Missing singular values are set to zero. The nullspace is local to an anchor:
multiple examples or snapshots need not leave a shared subspace of that
dimension. Support changes and truncation distinguish the geometric energies;
the $1/r$ reduction additionally changes the scale of the training loss.

\paragraph{Functional relevance.}
Let $g_s:\mathbb R^d\to\mathbb R^{c_s}$ be a frozen old readout,
$A_s=J_{g_s}(h_0)$, and let $V_r,\Sigma_r$ retain positive singular values only.
Writing $P_r=V_rV_r^\top$, define
$\alpha_{s,r}=\|A_sV_r\Sigma_r^{-1}\|_{\mathrm{op}}$ and
$\beta_{s,r}=\|A_s(I-P_r)\|_{\mathrm{op}}$. If $J_{g_s}$ is $L_g$-Lipschitz on a
convex neighborhood containing the displacement segment, then
\begin{equation}
\label{eq:functional_leakage_main}
    \|g_s(h_0+\delta)-g_s(h_0)\|_2
    \leq\alpha_{s,r}\sqrt{\delta^\top Q_r\delta}
       +\beta_{s,r}\|\delta\|_2+\frac{L_g}{2}\|\delta\|_2^2.
\end{equation}
This bound separates controlled geometric energy from old-readout sensitivity
outside the protected subspace. It is a conditional frozen-readout statement,
not a guarantee for the changing Class-IL classifier. We do not estimate
$\beta_{s,r}$ in the reported experiments. Appendix~\ref{app:math} provides the
proofs, support-boundary example, and further consequences.

\section{Experimental setup}

\textbf{Datasets.} CIFAR-100 (100 classes, split into 10 tasks of 10 classes
each; used both at native $32\times32$ resolution with a from scratch backbone,
and upsampled to $224\times224$ with a pretrained backbone). CUB-200 (200
fine-grained bird species, 10 tasks of 20 classes each, $224\times224$).
ImageNet-R (200 classes -- renditions/art styles of ImageNet classes, 10 tasks
of 20 classes each, $224\times224$). TinyImageNet (200 classes, split into 10
tasks of 20 classes each).
For the from scratch regime we use CIFAR-100 and TinyImageNet, which provide 500 training images per class, rather than CUB-200 and ImageNet-R (30 and 120 images per class), as training on such limited data is not possible. more details are described in Appendix~\ref{app:experimental_details}.

\textbf{Backbones.} ViT-B/16, pretrained on ImageNet-21k. 
ResNet-18, trained entirely from scratch. We include more
details on the hyperparameters in Appendix \ref{appsec:alt_hyperparams}

\textbf{Scenario and metrics.} All experiments use the class-incremental
learning (Class-IL) scenario, where the task identity is not available at test
time. Task-IL scenario (task identity given at test time) is reported alongside
as a secondary, less demanding diagnostic in Appendix \ref{appsec:taskil}. We implement our proposed methods within the Mammoth continual learning framework\footnote{\url{https://github.com/aimagelab/mammoth}} \cite{buzzega2020der,boschini2022class} and use it for all experiments. The primary metrics are final mean accuracy (accuracy on all classes seen so far, averaged over tasks, after training on the last task) and forgetting (mean drop from each task's peak accuracy to its accuracy at the end of training), computed using Mammoth's standard evaluation protocol.

\textbf{Protocol.} A replay buffer of size $500$ ($5120$ for TinyImagenet) is used throughout for every
rehearsal-based method (ours and all four baselines). Every number reported as
``final'' anywhere in this paper is the mean $\pm$ standard deviation over three
seeds ($\{0,1,2\}$) on the true test set; hyperparameter selection
(Section~\ref{appsec:alt_hyperparams}) is always performed separately on a
held-out validation split and never touches the test set used for the numbers
reported here.

\textbf{Anchor snapshots and memory.} The live reservoir contains $500$
examples, but \texttt{History} retains an additional frozen task-boundary
snapshot after each completed task. Each snapshot stores its then-current
reservoir images together with old representations and the corresponding anchor
factors. Thus History has auxiliary memory and anchor-computation costs that
grow with the number of completed tasks, even though the live replay reservoir
remains fixed at $500$ examples; we therefore do not claim a strictly
memory-matched comparison to ER, DER++, or ER-ACE. The default mean History
reduction in Eq.~\ref{eq:history_impl} keeps the nominal anchor-loss scale
approximately independent of the number of retained snapshots.

\textbf{Evaluation plan.} Our main experimental evaluation can be organized into
four parts: \textbf{(1)} Ablation regarding the design choices for the
task-boundary SAE-based anchoring and a comparison between SAE-Jacobian
anchoring and direct SAE-latent anchoring. In the ablation we include
combinations of \texttt{Vanilla Latent}/\texttt{Vanilla Jacobian},
\texttt{History} and \texttt{Replay} configurations on a full experimental stack
(Section \ref{sec:ablation}); \textbf{(2)} Comparison of the Jacobian-based
anchoring (\texttt{SAGA}) with established continual
learning baselines, including SGD (no forgetting mitigation), Experience Replay
(ER) \cite{chaudhry2019tiny,rolnick2019experience}, Dark Experience Replay with Knowledge
Distillation (DER++) \cite{buzzega2020der}, and Experience Replay with
Asymmetric Cross-Entropy (ER-ACE) \cite{caccia2021new} on a full experimental
stack (for further experimental details see Appendix
\ref{appsec:ref_geomteruies}, and for the results see Section
\ref{sec:comparison_with_std_techniques}); \textbf{(3)} Analysis of the effect
of the retained Jacobian rank (r) to assess performance in low-rank anchoring
regimes on example configurations (Section \ref{sec:rank_analysis}); \textbf{(4)}
Comparison of SAE-induced geometry with alternative reference subspaces used for
anchoring derived from random directions, Principal Component Analysis (PCA),
and classifier SVD on example experimental configuration (Section
\ref{sec:sanity_checks}). We also perform the analysis of the effect of SAE
dictionary size in Appendix \ref{sec:dic_size}, ranging from undercomplete to
strongly overcomplete configurations, to determine whether overcompleteness is
necessary for effective anchoring. For computational efficiency, we use
undercomplete SAEs in the main experiments, while our dictionary-size ablation
in Appendix \ref{sec:dic_size} shows that overcomplete SAEs achieve comparable
anchoring performance and can therefore be used when interpretability is also
desired.

\section{Experimental results}

\subsection{Comparison of Jacobian-based anchoring with Latent-based anchoring
with SAEs}\label{sec:ablation}

In this section, we compare \texttt{Jacobian} and direct \texttt{Latent}
anchoring separately for \textbf{pretrained} backbones (model: ViT-B/16,
datasets: CIFAR-100, CUB-200, ImageNet-R) and backbones trained \textbf{from
scratch} (model: ResNet-18, datasets: CIFAR-100, TinyImageNet), along with the
ablation study of the design choices for both these methods.
Tables~\ref{tab:ablation_pretrained} and~\ref{tab:ablation_scratch} report the
Class-IL accuracy and forgetting results for the pretrained and from scratch
cases respectively. Task-IL results are provided in
Appendix~\ref{appsec:taskil}.

\begin{table}[h]
\centering
\caption{SAE anchoring ablations for Class-IL scenario with \textbf{pretrained}
ViT-B/16. Results are mean $\pm$ std over three seeds. Higher accuracy
($\uparrow$) and lower forgetting ($\downarrow$) are better.}
\label{tab:ablation_pretrained}
\label{tab:forgetting_ablation_pretrained}
\resizebox{\linewidth}{!}{
\begin{tabular}{lcccccc}
\toprule
& \multicolumn{2}{c}{\textbf{CIFAR-100}}
& \multicolumn{2}{c}{\textbf{CUB-200}}
& \multicolumn{2}{c}{\textbf{ImageNet-R}} \\
\cmidrule(lr){2-3} \cmidrule(lr){4-5} \cmidrule(lr){6-7}
\textbf{Configuration}
& \textbf{Acc.} & \textbf{Forget.}
& \textbf{Acc.} & \textbf{Forget.}
& \textbf{Acc.} & \textbf{Forget.} \\
\midrule
Vanilla Latent
& $31.58 \pm 3.19$ & $72.95 \pm 3.64$
& $37.89 \pm 0.79$ & $51.36 \pm 0.70$
& $33.76 \pm 0.36$ & $64.69 \pm 0.25$ \\
Vanilla Jacobian
& $27.57 \pm 0.95$ & $78.69 \pm 1.13$
& $33.03 \pm 0.80$ & $60.94 \pm 0.46$
& $28.15 \pm 1.09$ & $71.86 \pm 1.28$ \\
\midrule
Latent + History
& $56.65 \pm 0.46$ & $45.77 \pm 0.54$
& $61.21 \pm 1.22$ & $27.77 \pm 1.26$
& $48.61 \pm 0.29$ & $49.33 \pm 0.24$ \\
Jacobian + History
& $51.37 \pm 1.89$ & $52.12 \pm 2.09$
& $59.53 \pm 0.37$ & $31.79 \pm 0.38$
& $45.06 \pm 1.47$ & $53.00 \pm 1.45$ \\
\midrule
Latent + Replay
& $70.96 \pm 0.21$ & $25.82 \pm 0.84$
& $64.45 \pm 1.64$ & $20.96 \pm 1.53$
& $53.78 \pm 0.19$ & $42.32 \pm 0.36$ \\
Jacobian + Replay
& $72.31 \pm 0.91$ & $28.49 \pm 1.15$
& $66.85 \pm 1.77$ & $23.04 \pm 1.79$
& $53.67 \pm 0.78$ & $43.38 \pm 0.98$ \\
\midrule
Latent + Replay + History
& $\mathbf{78.48 \pm 0.73}$ & $\mathbf{19.83 \pm 1.04}$
& $72.78 \pm 0.96$ & $\mathbf{14.04 \pm 0.92}$
& $58.96 \pm 1.18$ & $37.55 \pm 1.04$ \\
\textbf{SAGA (Jacobian + Replay + History)}
& $76.01 \pm 0.46$ & $24.24 \pm 0.43$
& $\mathbf{73.68 \pm 0.79}$ & $15.38 \pm 0.92$
& $\mathbf{61.58 \pm 0.51}$ & $\mathbf{34.48 \pm 0.40}$ \\
\bottomrule
\end{tabular}}
\end{table}

The clearest ablation result is the benefit of combining replay with historical
anchors. For both anchor types and all five settings, the full \texttt{Replay +
History} configuration improves both mean Class-IL accuracy and mean forgetting
over the corresponding \texttt{Replay}-only configuration. History alone helps
substantially on pretrained representations, but is less reliable from scratch.
Thus the useful setting is geometric anchoring together with replay and retained
snapshots, rather than either standalone anchor.

The comparison between the full latent and Jacobian configurations is more
specific. Jacobian anchoring improves mean accuracy in four of five settings.
On from scratch CIFAR-100, accuracy rises from $21.44\%$ to $28.53\%$ and
forgetting falls from $48.90$ to $38.46$; both metrics also improve on
TinyImageNet and pretrained ImageNet-R. CUB-200 trades slightly higher accuracy
for higher forgetting, while latent anchoring is better on CIFAR-100.
These outcomes characterize the implemented methods; the differing loss
reductions in Eqs.~\ref{eq:implemented_jacobian_loss} and
\eqref{eq:implemented_latent_loss} prevent attributing every difference solely
to the choice of local geometry.

\begin{table}[h]
\centering
\caption{SAE anchoring ablations for Class-IL scenario with ResNet-18 trained
\textbf{from scratch}. Results are mean $\pm$ std over three seeds. Higher
accuracy ($\uparrow$) and lower forgetting ($\downarrow$) are better.}
\label{tab:ablation_scratch}
\label{tab:forgetting_ablation_scratch}
\resizebox{\linewidth}{!}{
\begin{tabular}{lcccc}
\toprule
& \multicolumn{2}{c}{\textbf{CIFAR-100}}
& \multicolumn{2}{c}{\textbf{TinyImageNet}} \\
\cmidrule(lr){2-3} \cmidrule(lr){4-5}
\textbf{Configuration}
& \textbf{Acc.} & \textbf{Forget.}
& \textbf{Acc.} & \textbf{Forget.} \\
\midrule
Vanilla Latent
& $8.03 \pm 0.74$ & $59.32 \pm 3.53$
& $6.64 \pm 0.19$ & $57.79 \pm 3.37$ \\
Vanilla Jacobian
& $8.16 \pm 1.20$ & $69.50 \pm 5.43$
& $7.32 \pm 0.45$ & $61.78 \pm 4.50$ \\
\midrule
Latent + History
& $7.90 \pm 0.34$ & $58.14 \pm 4.18$
& $6.44 \pm 0.08$ & $58.85 \pm 3.26$ \\
Jacobian + History
& $8.20 \pm 0.52$ & $66.50 \pm 7.42$
& $5.89 \pm 0.62$ & $56.40 \pm 5.09$ \\
\midrule
Latent + Replay
& $15.78 \pm 0.29$ & $55.26 \pm 2.07$
& $17.79 \pm 0.83$ & $38.55 \pm 4.37$ \\
Jacobian + Replay
& $17.53 \pm 1.87$ & $55.24 \pm 3.69$
& $16.07 \pm 3.46$ & $40.24 \pm 5.92$ \\
\midrule
Latent + Replay + History
& $21.44 \pm 1.12$ & $48.90 \pm 0.96$
& $25.45 \pm 1.51$ & $17.10 \pm 0.62$ \\
\textbf{Jacobian + Replay + History}
& $\mathbf{28.53 \pm 0.52}$ & $\mathbf{38.46 \pm 2.64}$
& $\mathbf{28.27 \pm 4.43}$ & $\mathbf{12.84 \pm 2.24}$ \\
\bottomrule
\end{tabular}}
\end{table}

\subsection{Comparison between Jacobian-based SAE anchoring and common
forgetting mitigation methods}\label{sec:comparison_with_std_techniques}

In this section, we compare \texttt{Jacobian} anchoring with common continual
learning strategies -- SGD (no forgetting mitigation), ER, ER-ACE and DER++ --
separately for \textbf{pretrained} backbones (model: ViT-B/16, datasets:
CIFAR-100, CUB-200, ImageNet-R) and backbones trained \textbf{from scratch}
(model: ResNet-18, datasets: CIFAR-100, TinyImageNet).
Tables~\ref{tab:baseline_comparison_pretrained}
and~\ref{tab:baseline_comparison_scratch} report Class-IL accuracy and
forgetting for the pretrained and from scratch cases respectively. Task-IL
results are provided in Appendix~\ref{appsec:taskil}.

\begin{table}[h]
\centering
\caption{Class-IL accuracy and forgetting against continual learning baselines
on three \textbf{pretrained} ViT-B/16 settings. Results are mean $\pm$ std over
three seeds. Higher accuracy ($\uparrow$) and lower forgetting ($\downarrow$)
are better.}
\label{tab:baseline_comparison_pretrained}
\label{tab:forgetting_baselines_pretrained}
\scriptsize
\resizebox{\linewidth}{!}{
\begin{tabular}{lcccccc}
\toprule
& \multicolumn{2}{c}{\textbf{CIFAR-100}}
& \multicolumn{2}{c}{\textbf{CUB-200}}
& \multicolumn{2}{c}{\textbf{ImageNet-R}} \\
\cmidrule(lr){2-3} \cmidrule(lr){4-5} \cmidrule(lr){6-7}
\textbf{Method}
& \textbf{Acc.} & \textbf{Forget.}
& \textbf{Acc.} & \textbf{Forget.}
& \textbf{Acc.} & \textbf{Forget.} \\
\midrule
\textbf{SAGA (Jacobian + Replay + History)}
& $\mathbf{76.01 \pm 0.46}$ & $\mathbf{24.24 \pm 0.43}$
& $\mathbf{73.68 \pm 0.79}$ & $\mathbf{15.38 \pm 0.92}$
& $\mathbf{61.58 \pm 0.51}$ & $\mathbf{34.48 \pm 0.40}$ \\
DER++
& $57.18 \pm 2.52$ & $46.04 \pm 2.88$
& $68.89 \pm 3.17$ & $19.57 \pm 3.60$
& $58.36 \pm 2.09$ & $38.42 \pm 2.36$ \\
ER-ACE
& $63.86 \pm 3.34$ & $38.01 \pm 3.78$
& $56.93 \pm 1.79$ & $29.34 \pm 2.04$
& $57.09 \pm 1.50$ & $38.79 \pm 1.62$ \\
ER
& $45.61 \pm 2.54$ & $58.93 \pm 2.86$
& $64.22 \pm 3.17$ & $26.10 \pm 3.43$
& $52.85 \pm 1.22$ & $44.40 \pm 1.33$ \\
SGD
& $9.78 \pm 0.04$ & $93.77 \pm 0.15$
& $9.02 \pm 0.48$ & $77.48 \pm 0.89$
& $9.31 \pm 0.23$ & $81.94 \pm 0.18$ \\
\bottomrule
\end{tabular}}
\end{table}

The strongest result is consistent performance on the three pretrained
benchmarks. In Table~\ref{tab:baseline_comparison_pretrained}, our full method
has the highest mean accuracy and lowest mean forgetting among the conventional
baselines in every setting. Its Class-IL accuracy exceeds the strongest such
baseline by $12.15$ percentage points on CIFAR-100, $4.79$ on CUB-200, and
$3.22$ on ImageNet-R. These comparisons include the additional historical
snapshot storage described above; they are not equal-total-memory comparisons.

The from scratch results define the limits of this advantage
(Table~\ref{tab:baseline_comparison_scratch}). Our method improves on ER in
accuracy on both datasets, but DER++ or ER-ACE attains higher final accuracy.
On TinyImageNet it has the lowest mean forgetting, while on CIFAR-100 its
forgetting is similar to DER++ and ER-ACE. The contrast is consistent with
anchoring being particularly useful when adapting pretrained representations,
but architecture, data, and initialization differ between the regimes, therefore these
experiments do not isolate pretraining as the causal factor.

\begin{table}[h]
\centering
\caption{Class-IL accuracy and forgetting against continual learning baselines
with a ResNet-18 trained \textbf{from scratch}. Results are mean $\pm$ std over
three seeds. Higher accuracy ($\uparrow$) and lower forgetting ($\downarrow$)
are better.}
\label{tab:baseline_comparison_scratch}
\label{tab:forgetting_baselines_scratch}
\scriptsize
\begin{tabular}{lcccc}
\toprule
& \multicolumn{2}{c}{\textbf{CIFAR-100}}
& \multicolumn{2}{c}{\textbf{TinyImageNet}} \\
\cmidrule(lr){2-3} \cmidrule(lr){4-5}
\textbf{Method}
& \textbf{Acc.} & \textbf{Forget.}
& \textbf{Acc.} & \textbf{Forget.} \\
\midrule
\textbf{SAGA (Jacobian + Replay + History)}
& $28.53 \pm 0.52$ & $38.46 \pm 2.64$
& $28.27 \pm 4.43$ & $\mathbf{12.84 \pm 2.24}$ \\
DER++
& $\mathbf{36.28 \pm 0.24}$ & $37.71 \pm 1.20$
& $29.90 \pm 5.00$ & $37.61 \pm 3.96$ \\
ER-ACE
& $32.16 \pm 0.99$ & $\mathbf{37.41 \pm 3.88}$
& $\mathbf{32.37 \pm 0.49}$ & $29.98 \pm 2.04$ \\
ER
& $18.28 \pm 0.39$ & $71.89 \pm 0.64$
& $25.13 \pm 0.49$ & $52.09 \pm 0.34$ \\
SGD
& $8.37 \pm 0.71$ & $68.55 \pm 3.87$
& $6.84 \pm 0.72$ & $53.83 \pm 6.67$ \\
\bottomrule
\end{tabular}
\end{table}

\subsection{Effect of anchor rank $r$.}\label{sec:rank_analysis}

Table~\ref{tab:rank_ablation} studies the effect of the retained Jacobian rank $r\in{4,8,16,32,64}$ for SAGA (\texttt{Jacobian + Replay + History}), using a TopK SAE with $K=64$ and a pretrained ViT-B/16 backbone and three datasets. With raw weights, changing $r$ also changes the effective penalty strength because the loss averages over retained directions with decaying singular values. We therefore normalize each anchor's weights to unit mean square, preserving their relative weighting while keeping the nominal penalty strength constant across ranks.

Class-IL performance generally improves, however, substantially lower ranks remain competitive: even $r\ll K$ preserves strong performance, while Task-IL accuracy is comparatively insensitive to the retained rank. This indicates that the SAE-induced geometry admits substantial low-rank compression, although retaining a larger fraction of the Jacobian-sensitive subspace can be beneficial for Class-IL performance.


\begin{table}[h]
\centering
\caption{Effect of the retained Jacobian rank $r$ for a TopK SAE with $K=64$ on accuracy, with normalized anchor weights (constant nominal penalty strength across $r$) and pretrained ViT-B/16. Results are mean $\pm$ std over three seeds. Higher results are better ($\uparrow$).}
\label{tab:rank_ablation}
\scriptsize
\resizebox{\linewidth}{!}{
\begin{tabular}{lcccccc}
\toprule
& \multicolumn{2}{c}{\textbf{CIFAR-100}}
& \multicolumn{2}{c}{\textbf{ImageNet-R}}
& \multicolumn{2}{c}{\textbf{CUB-200}} \\
\cmidrule(lr){2-3} \cmidrule(lr){4-5} \cmidrule(lr){6-7}
\textbf{Configuration} & \textbf{Class-IL} & \textbf{Task-IL} & \textbf{Class-IL} & \textbf{Task-IL} & \textbf{Class-IL} & \textbf{Task-IL} \\
\midrule
Jacobian ($r=4$) + Replay + History & $75.55\pm0.56$ & $97.45\pm0.09$ & $59.41\pm0.35$ & $88.44\pm0.36$ & $70.25\pm1.35$ & $85.76\pm0.52$ \\
Jacobian ($r=8$) + Replay + History & $75.71\pm0.33$ & $97.43\pm0.17$ & $60.63\pm0.76$ & $88.55\pm0.48$ & $71.87\pm0.49$ & $85.92\pm0.67$ \\
Jacobian ($r=16$) + Replay + History & $76.48\pm0.28$ & $97.41\pm0.04$ & $61.64\pm0.63$ & $88.45\pm0.42$ & $74.06\pm0.82$ & $86.23\pm0.34$ \\
Jacobian ($r=32$) + Replay + History & $78.08\pm0.36$ & \textbf{$97.50\pm0.02$} & $63.94\pm0.16$ & $88.87\pm0.24$ & $74.52\pm0.61$ & \textbf{$86.30\pm0.27$} \\
Jacobian ($r=64$) + Replay + History & \textbf{$79.45\pm0.11$} & $97.43\pm0.13$ & \textbf{$64.34\pm0.56$} & \textbf{$89.02\pm0.10$} & \textbf{$74.77\pm0.34$} & $86.01\pm0.74$ \\
\bottomrule
\end{tabular}}
\end{table}


\subsection{Comparison of Low-Rank Reference Geometries}\label{sec:sanity_checks}

To disentangle the contribution of the SAE-induced geometry from low-rank
anchoring more broadly, we compare SAE-Jacobian anchoring with random, PCA, and
classifier SVD reference \emph{geometries} of matched nominal rank for
pretrained Class-IL settings (Table~\ref{tab:geometry_pretrained}) and a
from scratch setting (Table~\ref{tab:geometry_rank_sweep}). All rows use
normalized anchor weighting, which fixes the total trace of the quadratic metric
while retaining each geometry's protected directions and relative
within-subspace spectral weights (Appendix~\ref{appsec:ref_geomteruies}). Thus
PCA and classifier SVD comparisons vary more than the subspace alone, whereas
the SAE geometry is additionally input dependent.

The normalized controls show that the choice of geometry matters, but not in a
uniform way. SAE-Jacobian has the highest mean accuracy and lowest mean
forgetting on CUB-200. On ImageNet-R, it is close to PCA and random anchoring:
the $0.07$-point accuracy difference from PCA is small relative to the reported
standard deviations. On the other hand, classifier SVD performs best on CIFAR-100. Therefore, these
results support local SAE geometry as a useful option, not a universal
replacement for global subspace constructions.

\begin{table}[h]
\centering
\caption{Class-IL accuracy and forgetting for different reference geometries on \textbf{pretrained} ViT-B/16. All variants use normalized anchor weighting. Results are mean $\pm$ std over three seeds. Higher accuracy ($\uparrow$) and lower forgetting ($\downarrow$) are better.}
\label{tab:geometry_pretrained}
\scriptsize
\resizebox{\linewidth}{!}{
\begin{tabular}{lcccccc}
\toprule
& \multicolumn{2}{c}{\textbf{CUB-200 ($r=16$)}}
& \multicolumn{2}{c}{\textbf{ImageNet-R ($r=16$)}}
& \multicolumn{2}{c}{\textbf{CIFAR-100 ($r=8$)}} \\
\cmidrule(lr){2-3} \cmidrule(lr){4-5} \cmidrule(lr){6-7}
\textbf{Reference}
& \textbf{Acc.} & \textbf{Forget.}
& \textbf{Acc.} & \textbf{Forget.}
& \textbf{Acc.} & \textbf{Forget.} \\
\midrule
Random
& $70.68 \pm 0.52$ & $19.05 \pm 0.84$
& $61.11 \pm 0.80$ & $34.78 \pm 1.03$
& $71.69 \pm 1.07$ & $29.11 \pm 1.12$ \\
PCA
& $71.64 \pm 0.92$ & $17.71 \pm 1.12$
& $61.37 \pm 0.25$ & $34.95 \pm 0.30$
& $74.98 \pm 0.55$ & $25.45 \pm 0.62$ \\
Classifier SVD
& $67.27 \pm 1.54$ & $22.66 \pm 1.67$
& $56.69 \pm 0.25$ & $39.81 \pm 0.52$
& $\mathbf{77.34 \pm 1.01}$ & $\mathbf{22.73 \pm 0.97}$ \\
SAE Jacobian
& $\mathbf{72.55 \pm 1.19}$ & $\mathbf{16.64 \pm 1.40}$
& $\mathbf{61.44 \pm 0.30}$ & $\mathbf{34.67 \pm 0.21}$
& $76.08 \pm 0.64$ & $24.16 \pm 0.75$ \\
\bottomrule
\end{tabular}}
\end{table}

For the from scratch setting (Tab. \ref{tab:geometry_rank_sweep}), we additionally vary the nominal retained rank
across all reference geometries. SAE-Jacobian is particularly effective at $r=4$
and $r=8$ in Class-IL accuracy, while PCA has lower mean forgetting at every
displayed rank. At $r=16$, PCA and SAE-Jacobian have nearly identical mean
accuracy, and at $r=64$ SAE-Jacobian again has higher mean accuracy. These
results show that alternative geometries occupy different stability--plasticity
operating points rather than supporting a universal ordering. The normalized
comparison controls the total metric trace, but PCA, classifier SVD, and
SAE-Jacobian retain their own relative spectral weighting; the experiment is
therefore a comparison of low-rank geometries rather than of directions alone.
The functional relevance of the SAE geometry is characterized by the leakage
term $\beta_{s,r}$ in Eq.~\ref{eq:functional_leakage_main}, which describes
old-readout sensitivity outside the protected subspace.

\begin{table}[h]
\centering
\caption{Class-IL accuracy and forgetting for different reference geometries and
ranks on CIFAR-100 with ResNet-18 trained \textbf{from scratch} ($r=K$). All
variants use normalized anchor weighting. Results are mean $\pm$ std over three
seeds. Higher accuracy ($\uparrow$) and lower forgetting ($\downarrow$) are
better.}
\label{tab:geometry_rank_sweep}
\resizebox{\linewidth}{!}{
\begin{tabular}{lcccccccc}
\toprule
& \multicolumn{2}{c}{\textbf{Random}}
& \multicolumn{2}{c}{\textbf{PCA}}
& \multicolumn{2}{c}{\textbf{Classifier SVD}}
& \multicolumn{2}{c}{\textbf{SAE Jacobian}} \\
\cmidrule(lr){2-3} \cmidrule(lr){4-5} \cmidrule(lr){6-7} \cmidrule(lr){8-9}
\textbf{Rank $r$}
& \textbf{Acc.} & \textbf{Forget.}
& \textbf{Acc.} & \textbf{Forget.}
& \textbf{Acc.} & \textbf{Forget.}
& \textbf{Acc.} & \textbf{Forget.} \\
\midrule
4
& $20.83 \pm 0.39$ & $51.76 \pm 2.74$
& $19.81 \pm 2.30$ & $\mathbf{32.57 \pm 7.77}$
& $21.99 \pm 1.55$ & $42.86 \pm 6.33$
& $\mathbf{23.63 \pm 0.71}$ & $46.65 \pm 3.06$ \\
8
& $20.69 \pm 3.96$ & $46.23 \pm 5.20$
& $22.48 \pm 0.95$ & $\mathbf{36.46 \pm 7.52}$
& $22.52 \pm 1.26$ & $43.63 \pm 8.14$
& $\mathbf{24.88 \pm 1.85}$ & $37.74 \pm 2.69$ \\
16
& $23.39 \pm 1.07$ & $45.26 \pm 2.44$
& $\mathbf{27.71 \pm 2.21}$ & $\mathbf{27.62 \pm 7.20}$
& $26.29 \pm 1.33$ & $41.92 \pm 1.19$
& $27.65 \pm 1.44$ & $35.44 \pm 4.94$ \\
64
& $23.22 \pm 1.72$ & $45.73 \pm 1.71$
& $19.07 \pm 8.06$ & $\mathbf{28.81 \pm 6.75}$
& $24.39 \pm 5.52$ & $40.52 \pm 5.01$
& $\mathbf{26.19 \pm 4.58}$ & $33.03 \pm 4.27$ \\
\bottomrule
\end{tabular}}
\end{table}

\section{Conclusions}

We introduced Sparse Autoencoder Geometric Anchoring (SAGA), a representation-level continual learning regularizer in which task-specific sparse autoencoders induce local positive-semidefinite geometries over stored representations. Replay snapshots specify where anchoring is applied, while truncated Jacobian singular directions determine which local changes are constrained. Our results show that SAEs can serve not only as latent feature spaces, but also as sources of sample-specific geometric constraints on backbone drift. These constraints can be compressed to a chosen rank and combined with replay supervision.
The strongest empirical result is the full-stack SAGA's accuracy and forgetting
relative to conventional baselines across all three pretrained benchmarks.
From scratch and geometry control results show that the benefit is regime dependent, while the rank ablation shows that the SAE-induced geometry can be substantially compressed with limited performance loss. The contribution, therefore, is a concrete use of
SAEs as local geometric memory, together with a theoretical analysis of what
that memory controls and an empirical characterization of where it helps.


\paragraph{Limitations.}
The Jacobian is a local linear description of the SAE and, for hard TopK
encoders, its exact relation to latent anchoring holds only within a fixed
activation/support cell; support transitions are a genuine source of
discrepancy. An unsupervised SAE also does not guarantee that its sensitive
directions coincide with old-task functional sensitivity, which is why our
theoretical bound contains an explicit leakage term and why the
PCA/classifier/random controls are important. 
Results use three seeds, so small differences in
means should not be over-interpreted. Historical anchors also retain complete
task-boundary reservoir snapshots and their stored representations/geometric
factors, so their auxiliary memory and compute grow with the number of completed
tasks even though the live replay reservoir remains fixed at $500$ examples.
Future work should therefore investigate anchor compression, more efficient
SAE/Jacobian construction, and the resulting memory--accuracy--computation
trade-offs.

\FloatBarrier

\subsection*{AI use statement}

Generative AI tools were used to assist with research ideation, mathematical
checking, drafting, and document organization. The authors reviewed the
AI-assisted material and independently verified the mathematical statements,
experimental results, citations, implementation decisions, and final claims
included in the submission. The authors take responsibility for the final
content of the work. Generative AI also was used to rewrite handwritten formulas
and whiteboard photos into LaTeX equations.

\subsection*{Ethics statement}

This work does not involve any sensitive aspects, such as human subjects,
personal data, or user studies. Continual learning is an important enabler for
developing adaptive machine learning systems that can incorporate new knowledge
while preserving previously acquired information. Our work aims to advance this
area by addressing catastrophic forgetting -- its key challenge -- through an
anchoring method. In conducting and reporting the study, we aimed to uphold high
standards of scientific excellence through transparent methodology, reproducible
experimental procedures, sharing our code, making it compatible with a commonly
used framework (Mammoth) and careful reporting of both strengths and
limitations. We are not aware of direct harmful applications specific to the
proposed method. Nevertheless, as with other machine learning methods, its use
in deployed systems may inherit risks associated with the underlying application
domains, and such downstream effects should be considered.

\subsection*{Reproducibility Statement}
To support reproducibility, we evaluated our method on  publicly available continual learning datasets and implement it within the widely used Mammoth continual learning framework. Our implementation follows the experimental settings and evaluation protocols described in the paper, and we provide the source code required to reproduce the experiments. The code will be made publicly available on GitHub upon acceptance.


\subsubsection*{Acknowledgments}

This work was supported by the National Science Centre, Poland, under the OPUS grant 2025/59/B/ST6/00193.

\bibliography{iclr2027_conference}
\bibliographystyle{iclr2027_conference}

\appendix

\section{Mathematical details of task-anchored SAE geometry}
\label{app:math}

\subsection{Local pullback approximation}
For a frozen task-boundary SAE encoder $E_s$, define the nonlinear displacement
energy at an anchor $h_0$ by
\begin{equation}
    d_s(h_0,h_0+\delta)^2
    =\|E_s(h_0+\delta)-E_s(h_0)\|_2^2 .
\end{equation}
Assume that $E_s$ is differentiable near $h_0$ and that its Jacobian is
$L_s$-Lipschitz in operator norm. Writing $J=J_{E_s}(h_0)$, Taylor's theorem
gives
\begin{equation}
    E_s(h_0+\delta)-E_s(h_0)=J\delta+r(\delta),
    \qquad
    \|r(\delta)\|_2\leq \frac{L_s}{2}\|\delta\|_2^2 .
\end{equation}
Therefore
\begin{align}
\left|
    d_s(h_0,h_0+\delta)^2
    -\delta^\top J^\top J\delta
\right|
&\leq
    L_s\|J\|_{\mathrm{op}}\|\delta\|_2^3
    +\frac{L_s^2}{4}\|\delta\|_2^4 .
\end{align}
Thus $J^\top J$ is the local pullback semimetric induced by Euclidean distance
in SAE latent space.

\subsection{Exact affine ReLU--TopK geometry}
Assume
\begin{equation}
    E(h)=\mathrm{TopK}(\mathrm{ReLU}(Wh+b))
\end{equation}
where $\mathrm{TopK}$ is the hard coordinate-masking operator used in the
implementation: it keeps the $K$ largest post-ReLU values in their original
latent coordinates and sets all remaining coordinates to zero. Let $h_0$ be
a regular point whose selected TopK coordinates are strictly positive and
separated from the remaining coordinates. Let $S(h_0)$ be the selected support
and let $M_{S(h_0)}$ denote the corresponding diagonal mask. Then
\begin{equation}
    J_E(h_0)=M_{S(h_0)}W.
\end{equation}
Consequently, for every displacement $\delta$,
\begin{equation}
    \delta^\top J_E(h_0)^\top J_E(h_0)\delta
    =\|M_{S(h_0)}W\delta\|_2^2,
\end{equation}
so full frozen-Jacobian anchoring is exactly matching the old active
preactivation changes under the frozen anchor mask.

If the ReLU/TopK mask remains unchanged on the segment
$\{h_0+\tau\delta:0\leq\tau\leq1\}$, the encoder is affine on that segment and
hence
\begin{equation}
    E(h_0+\delta)-E(h_0)=J_E(h_0)\delta.
\end{equation}
Therefore the \emph{unnormalized geometric energy} of the full Jacobian equals
the SAE-latent displacement exactly inside one cell. Because
$\mathrm{rank}(J_E(h_0))\leq K$, retaining every nonzero singular direction also
gives equality of these per-example energies. The training reductions are not
identical: the implementation divides the Jacobian energy by the retained rank
$r$ through its mean over directions, whereas the latent loss sums
latent-coordinate errors. Hence fixed-cell equality should not be read as
equality of the two implemented losses at the same numerical anchor weight.

A simple sufficient support-stability certificate can be written in terms of the
TopK gap. Let
\begin{equation}
    z(h)=\mathrm{ReLU}(Wh+b)
\end{equation}
and let $z_{(1)}(h_0)\geq\cdots\geq z_{(m)}(h_0)$ be the ordered coordinates. For $K<m$, define
\begin{equation}
    \gamma_K(h_0)=z_{(K)}(h_0)-z_{(K+1)}(h_0).
\end{equation}
If $\gamma_K(h_0)>0$ and
\begin{equation}
    \|W\delta\|_\infty<\frac{\gamma_K(h_0)}{2},
\end{equation}
then the TopK support is unchanged. Since coordinatewise ReLU is 1-Lipschitz,
every post-ReLU coordinate moves by less than $\gamma_K/2$, so the ordering
across the $K/(K+1)$ gap cannot reverse. A sufficient Euclidean condition is
\begin{equation}
    \|\delta\|_2<
    \frac{\gamma_K(h_0)}{2\|W\|_{2\to\infty}},
    \qquad
    \|W\|_{2\to\infty}=\max_j\|W_{j,:}\|_2.
\end{equation}

Support changes can nevertheless separate latent and frozen-Jacobian anchoring
sharply. For $K=1$, $W=I_2$, $b=0$, $a>0$,
\begin{equation}
    h_0=(a+\varepsilon,a),
    \qquad
    h_1=(a,a+\varepsilon),
\end{equation}
we have $\|h_1-h_0\|_2=\sqrt{2}\,\varepsilon\to0$, while
\begin{equation}
    \|E(h_1)-E(h_0)\|_2^2=2(a+\varepsilon)^2\to2a^2,
\end{equation}
but the frozen full-Jacobian penalty at $h_0$ is only $\varepsilon^2\to0$. Thus
hard TopK support transitions are one mechanism by which latent matching and
frozen local geometry genuinely differ even without rank truncation.

\subsection{Optimal rank truncation}
Let
\begin{equation}
    Q=J^\top J=\sum_{j=1}^q\sigma_j^2v_jv_j^\top,
    \qquad
    \sigma_1\geq\cdots\geq\sigma_q>0,
\end{equation}
For $1\le r\le q$, set $\sigma_{q+1}=0$ and define
\begin{equation}
    Q_r=\sum_{j=1}^r\sigma_j^2v_jv_j^\top.
\end{equation}
By the Eckart--Young--Mirsky theorem, $Q_r$ is a best rank-$r$ approximation to
$Q$ in both operator and Frobenius norms, with
\begin{equation}
    \|Q-Q_r\|_{\mathrm{op}}=\sigma_{r+1}^2,
    \qquad
    \|Q-Q_r\|_F^2=\sum_{j>r}\sigma_j^4.
\end{equation}
For every $\delta$,
\begin{equation}
    0\leq\delta^\top(Q-Q_r)\delta
    =\sum_{j>r}\sigma_j^2(v_j^\top\delta)^2
    \leq\sigma_{r+1}^2\|\delta\|_2^2.
\end{equation}
Moreover, $\mathrm{rank}(Q_r)\leq r$, so $\dim\ker Q_r\geq d-r$.

\subsection{Spatial sampling and directional compression}
Let $\mu_s^{\mathrm{snap}}$ denote the distribution of input examples in the cumulative stream at task boundary $s$ that the reservoir snapshot $B_s$ is intended
to approximate. Define the ideal nonlinear snapshot-anchor energy
\begin{equation}
    R_s^{\mathrm{SAE}}(\theta)
    =\mathbb{E}_{x\sim\mu_s^{\mathrm{snap}}}
    \left[\|E_s(f_\theta(x))-E_s(h_s(x))\|_2^2\right].
\end{equation}
With $\delta_s(x;\theta)=f_\theta(x)-h_s(x)$, set
\begin{equation}
    q_s(x;\theta)=\delta_s^\top Q_s(h_s(x))\delta_s,
    \qquad
    q_{s,r}(x;\theta)=\delta_s^\top Q_{s,r}(x)\delta_s,
\end{equation}
where $Q_s$ is the full pullback metric from Eq.~\ref{eq:full_Qs}. Then
\begin{align}
\left|R_s^{\mathrm{SAE}}(\theta)-\widehat R_{s,r}(\theta)\right|
\leq {}&
\underbrace{\mathbb{E}_{x\sim\mu_s^{\mathrm{snap}}}
\left|\|E_s(h_s+\delta_s)-E_s(h_s)\|_2^2-q_s(x;\theta)\right|}_{\mathcal{E}_{\mathrm{local}}}
\\
&+\underbrace{\mathbb{E}_{x\sim\mu_s^{\mathrm{snap}}}
\left[q_s(x;\theta)-q_{s,r}(x;\theta)\right]}_{\mathcal{E}_{\mathrm{rank}}}
\\
&+\underbrace{\left|\mathbb{E}_{x\sim\mu_s^{\mathrm{snap}}}q_{s,r}(x;\theta)
-\frac{1}{|B_s|}\sum_{x_i\in B_s}q_{s,r}(x_i;\theta)\right|}_{\mathcal{E}_{\mathrm{sample}}}.
\end{align}
The three terms quantify local/support error, directional compression, and finite spatial sampling. The rank term obeys
\begin{equation}
    \mathcal{E}_{\mathrm{rank}}
    \leq\mathbb{E}_{x\sim\mu_s^{\mathrm{snap}}}
    \left[\sigma_{s,r+1}(x)^2\|\delta_s(x;\theta)\|_2^2\right],
\end{equation}
and $\mathcal{E}_{\mathrm{local}}=0$ inside a fixed affine ReLU--TopK cell. This
decomposition is written for the unnormalized geometric energy $\widehat
R_{s,r}$; the raw Jacobian training loss carries the additional factor $1/r$
from Eq.~\ref{eq:implemented_jacobian_loss}.

For intuition about finite spatial resolution, fix the entire loss function,
including $h_s$, $E_s$, $r$, and $\theta$, independently of a hypothetical
i.i.d. evaluation sample $B_s$ from $\mu_s^{\mathrm{snap}}$. If $0\leq
q_{s,r}(x;\theta)\leq M_s$ almost surely, Hoeffding's inequality gives, with
probability at least $1-\eta$,
\begin{equation}
    \mathcal{E}_{\mathrm{sample}}
    \leq M_s\sqrt{\frac{\log(2/\eta)}{2|B_s|}}.
\end{equation}
This is an illustrative fixed-model sampling statement, not a generalization
guarantee for the trained continual learner: the implementation uses reservoir
snapshots to fit the SAE and later optimize the backbone, so the independence
assumption does not hold for those same stored examples.

\subsection{Functional leakage outside the protected subspace}
Let $g_s:\mathbb{R}^d\to\mathbb{R}^{c_s}$ be a frozen old-task logit map and let $A_s=J_{g_s}(h_0)$. Let $Q_r=V_r\Sigma_r^2V_r^\top$ be formed from $r$ retained \emph{nonzero} singular values, with $\sigma_r>0$, and set $P_r=V_rV_r^\top$. Define
\begin{equation}
    \alpha_{s,r}=\|A_sV_r\Sigma_r^{-1}\|_{\mathrm{op}},
    \qquad
    \beta_{s,r}=\|A_s(I-P_r)\|_{\mathrm{op}}.
\end{equation}
If $g_s$ has an $L_g$-Lipschitz Jacobian on a convex neighborhood containing
the displacement segment, Taylor expansion gives
\begin{equation}
    g_s(h_0+\delta)-g_s(h_0)=A_s\delta+\rho(\delta),
    \qquad
    \|\rho(\delta)\|_2\leq\frac{L_g}{2}\|\delta\|_2^2.
\end{equation}
Decomposing $A_s\delta=A_sP_r\delta+A_s(I-P_r)\delta$ and using
\begin{equation}
    A_sP_r\delta
    =A_sV_r\Sigma_r^{-1}(\Sigma_rV_r^\top\delta)
\end{equation}
yields
\begin{equation}
    \|g_s(h_0+\delta)-g_s(h_0)\|_2
    \leq
    \alpha_{s,r}\sqrt{\delta^\top Q_r\delta}
    +\beta_{s,r}\|\delta\|_2
    +\frac{L_g}{2}\|\delta\|_2^2.
\end{equation}
Thus a low-rank SAE anchor can control old-task function only to the extent that
old-task sensitivity outside the protected subspace, described by $\beta_{s,r}$,
is small.

This statement concerns compatibility with the frozen old-task readout. In the
Class-IL experiments the live classifier continues to change, so
classifier-parameter drift and competition from newly introduced classes can
alter logits even when the representation remains compatible with the old frozen
readout.

For a linear old readout $g_s(h)=A_sh+b_s$ and any $Q\succeq0$, there exists a
finite $c$ such that
\begin{equation}
    \|A_s\delta\|_2^2\leq c\,\delta^\top Q\delta
    \qquad\text{for all }\delta
\end{equation}
if and only if
\begin{equation}
    \ker Q\subseteq\ker A_s.
\end{equation}
When this holds, the smallest constant is
\begin{equation}
    c_\star=\|A_sQ^{\dagger/2}\|_{\mathrm{op}}^2.
\end{equation}
This gives an exact algebraic version of the same requirement: no
old-task-sensitive direction may lie completely in the anchor nullspace.

\subsection{Same-point historical quadratic aggregation}
The following identity applies when several historical quadratic anchors constrain the \emph{same} current representation $h$. It is useful for interpreting how multiple local PSD constraints combine, but it is not a literal reduction of the full implementation, where different task-boundary snapshots generally contain different replay examples and are evaluated on independently sampled minibatches. Let
\begin{equation}
    L_{\mathrm{hist}}(h)
    =\sum_{s=1}^T\omega_s(h-h_s)^\top Q_s(h-h_s),
    \qquad \omega_s>0,
\end{equation}
with $Q_s\succeq0$. Define
\begin{equation}
    Q_H=\sum_s\omega_sQ_s,
    \qquad
    b_H=\sum_s\omega_sQ_sh_s,
    \qquad
    c_H=\sum_s\omega_sh_s^\top Q_sh_s.
\end{equation}
With $h_\star=Q_H^\dagger b_H$,
\begin{equation}
    L_{\mathrm{hist}}(h)
    =(h-h_\star)^\top Q_H(h-h_\star)+\kappa,
\end{equation}
where $\kappa=c_H-b_H^\top Q_H^\dagger b_H\geq0$. Since all $Q_s$ are positive
semidefinite,
\begin{equation}
    \ker Q_H=\bigcap_{s=1}^T\ker Q_s.
\end{equation}
Thus, at a common point, adding PSD historical anchors can only shrink the
common unconstrained subspace. In the actual training procedure, History instead
averages or sums stochastic losses from separately stored task-boundary
snapshots; the default is the mean reduction, so total nominal anchor-loss scale
does not grow linearly with the number of snapshots, although storage and
compute do.

\section{Task-IL results}\label{appsec:taskil} For completeness, we report
Task-IL accuracy and forgetting corresponding to the main Class-IL experiments.
Task-IL assumes that the task identity is available at test time and serves as a
secondary diagnostic.

\begin{table}[h]
\centering
\caption{Task-IL accuracy and forgetting for SAE anchoring ablations with a pretrained ViT-B/16. Results are mean $\pm$ std over three seeds.}
\resizebox{\linewidth}{!}{
\begin{tabular}{lcccccc}
\toprule
& \multicolumn{2}{c}{\textbf{CIFAR-100}}
& \multicolumn{2}{c}{\textbf{CUB-200}}
& \multicolumn{2}{c}{\textbf{ImageNet-R}} \\
\cmidrule(lr){2-3}\cmidrule(lr){4-5}\cmidrule(lr){6-7}
\textbf{Configuration} & \textbf{Acc.} & \textbf{Forget.} & \textbf{Acc.} & \textbf{Forget.} & \textbf{Acc.} & \textbf{Forget.}\\
\midrule
Vanilla Latent & $89.34\pm0.30$ & $9.07\pm0.42$ & $74.97\pm0.29$ & $10.30\pm0.32$ & $84.29\pm0.62$ & $8.65\pm0.45$\\
Vanilla Jacobian & $93.10\pm0.59$ & $5.96\pm0.67$ & $79.23\pm1.49$ & $9.62\pm1.38$ & $84.00\pm0.53$ & $9.83\pm0.65$\\
\midrule
Latent + History & $95.66\pm0.12$ & $2.79\pm0.15$ & $82.28\pm0.81$ & $4.61\pm0.83$ & $87.33\pm0.60$ & $6.41\pm0.45$\\
Jacobian + History & $96.57\pm0.13$ & $2.00\pm0.12$ & $84.21\pm0.22$ & $4.49\pm0.26$ & $87.30\pm0.22$ & $6.15\pm0.53$\\
\midrule
Latent + Replay & $94.59\pm0.37$ & $3.24\pm0.37$ & $81.01\pm0.43$ & $4.35\pm0.53$ & $86.66\pm0.28$ & $6.27\pm0.14$\\
Jacobian + Replay & $97.17\pm0.03$ & $1.53\pm0.06$ & $84.56\pm0.75$ & $4.08\pm0.68$ & $87.24\pm0.20$ & $6.37\pm0.32$\\
\midrule
Latent + Replay + History & $97.24\pm0.12$ & $1.10\pm0.17$ & $84.85\pm0.78$ & $\mathbf{2.30\pm0.59}$ & $\mathbf{89.00\pm0.25}$ & $\mathbf{4.66\pm0.02}$\\
SAGA (Jacobian + Replay + History) & $\mathbf{97.54\pm0.10}$ & $\mathbf{1.03\pm0.18}$ & $\mathbf{86.38\pm0.34}$ & $2.42\pm0.47$ & $88.76\pm0.39$ & $\mathbf{4.66\pm0.23}$\\
\bottomrule
\end{tabular}}
\end{table}

\begin{table}[h]
\centering
\caption{Task-IL accuracy and forgetting for SAE anchoring ablations with a ResNet-18 trained from scratch.}
\scriptsize
\begin{tabular}{lcccc}
\toprule
& \multicolumn{2}{c}{\textbf{CIFAR-100}} & \multicolumn{2}{c}{\textbf{TinyImageNet}}\\
\cmidrule(lr){2-3}\cmidrule(lr){4-5}
\textbf{Configuration} & \textbf{Acc.} & \textbf{Forget.} & \textbf{Acc.} & \textbf{Forget.}\\
\midrule
Vanilla Latent & $59.03\pm4.49$ & $9.12\pm2.16$ & $60.86\pm1.28$ & $1.78\pm0.24$\\
Vanilla Jacobian & $54.51\pm8.59$ & $21.82\pm3.10$ & $47.29\pm5.19$ & $22.41\pm2.19$\\
\midrule
Latent + History & $65.88\pm2.20$ & $2.89\pm0.63$ & $61.34\pm0.66$ & $2.01\pm0.19$\\
Jacobian + History & $68.50\pm5.34$ & $5.79\pm0.92$ & $59.19\pm3.94$ & $3.47\pm0.44$\\
\midrule
Latent + Replay & $67.19\pm1.47$ & $9.21\pm2.46$ & $56.24\pm2.06$ & $7.94\pm1.43$\\
Jacobian + Replay & $68.50\pm1.64$ & $11.25\pm2.02$ & $51.98\pm6.02$ & $16.55\pm4.48$\\
\midrule
Latent + Replay + History & $72.16\pm0.66$ & $3.97\pm0.66$ & $58.61\pm1.46$ & $\mathbf{2.31\pm0.41}$\\
SAGA (Jacobian + Replay + History) & $\mathbf{74.35\pm0.43}$ & $\mathbf{3.93\pm0.90}$ & $\mathbf{61.84\pm3.38}$ & $3.04\pm1.34$\\
\bottomrule
\end{tabular}
\end{table}

Overall, the Task-IL results are consistent with the Class-IL findings included
in the main paper. In particular, combining replay with persistent task history
remains the strongest SAE-anchoring configuration, while SAE-Jacobian anchoring
shows particularly strong performance in the pretrained setting, as well
compared to the SAE latent anchoring in the from scratch scenario. Our method
also consistently reduces forgetting relative to standard baselines. 

\begin{table}[h]
\centering
\caption{Task-IL accuracy and forgetting against continual learning baselines on pretrained ViT-B/16.}
\resizebox{\linewidth}{!}{
\begin{tabular}{lcccccc}
\toprule
& \multicolumn{2}{c}{\textbf{CIFAR-100}} & \multicolumn{2}{c}{\textbf{CUB-200}} & \multicolumn{2}{c}{\textbf{ImageNet-R}}\\
\cmidrule(lr){2-3}\cmidrule(lr){4-5}\cmidrule(lr){6-7}
\textbf{Method} & \textbf{Acc.} & \textbf{Forget.} & \textbf{Acc.} & \textbf{Forget.} & \textbf{Acc.} & \textbf{Forget.}\\
\midrule
Ours & $\mathbf{97.54\pm0.10}$ & $\mathbf{1.03\pm0.18}$ & $\mathbf{86.38\pm0.34}$ & $\mathbf{2.42\pm0.47}$ & $\mathbf{88.76\pm0.39}$ & $\mathbf{4.66\pm0.23}$\\
DER++ & $91.64\pm0.91$ & $7.90\pm0.87$ & $83.14\pm0.98$ & $5.30\pm1.03$ & $85.77\pm0.49$ & $8.42\pm0.53$\\
ER-ACE & $93.90\pm1.10$ & $5.11\pm1.24$ & $77.92\pm1.46$ & $7.83\pm1.65$ & $84.10\pm0.38$ & $9.97\pm0.40$\\
ER & $95.16\pm0.33$ & $3.94\pm0.39$ & $83.01\pm0.75$ & $5.76\pm0.79$ & $86.00\pm0.58$ & $7.82\pm0.68$\\
SGD & $74.31\pm1.13$ & $22.07\pm1.37$ & $39.48\pm1.03$ & $43.63\pm1.21$ & $53.57\pm1.96$ & $32.78\pm2.16$\\
\bottomrule
\end{tabular}}
\end{table}

\begin{table}[h]
\centering
\caption{Task-IL accuracy and forgetting against continual learning baselines with a ResNet-18 trained from scratch.}
\scriptsize
\begin{tabular}{lcccc}
\toprule
& \multicolumn{2}{c}{\textbf{CIFAR-100}} & \multicolumn{2}{c}{\textbf{TinyImageNet}}\\
\cmidrule(lr){2-3}\cmidrule(lr){4-5}
\textbf{Method} & \textbf{Acc.} & \textbf{Forget.} & \textbf{Acc.} & \textbf{Forget.}\\
\midrule
Ours & $74.35\pm0.43$ & $\mathbf{3.93\pm0.90}$ & $61.84\pm3.38$ & $\mathbf{3.04\pm1.34}$\\
DER++ & $\mathbf{75.15\pm0.24}$ & $8.03\pm1.14$ & $66.08\pm4.69$ & $8.81\pm3.47$\\
ER-ACE & $70.00\pm0.56$ & $12.34\pm3.40$ & $61.10\pm0.55$ & $12.88\pm1.09$\\
ER & $68.13\pm0.86$ & $17.28\pm0.71$ & $\mathbf{66.09\pm0.31}$ & $9.65\pm0.44$\\
SGD & $32.49\pm3.01$ & $41.89\pm2.35$ & $17.13\pm2.25$ & $42.96\pm5.99$\\
\bottomrule
\end{tabular}
\end{table}


\FloatBarrier

\section{Effect of SAE dictionary size
(overcompleteness)}\label{sec:dic_size}

We perform all of our main experiments with SAE dictionary of size $256$, i.e.\
an \emph{undercomplete} dictionary relative to the ResNet-18 and ViT feature
dimensions ($D=512$, an expansion factor of $0.5\times$), whereas SAEs are more
commonly trained overcomplete when used for representational disentanglement and
interpretability. Table~\ref{tab:latentdim_ablation} tests different dictionary
sizes on CIFAR-100 with ResNet-18, ranging from $0.5\times$ (the default for
dictionary size $256$ in the ResNet-18 case) up to $8\times$, at fixed
$\texttt{sae\_topk}=\texttt{jacobian\_k}=16$. Class-IL accuracy is rather flat
across dictionary sizes and within one standard deviation of the default
$256$-unit dictionary at every size tested. The only configuration that is both
worse and markedly less stable across seeds is dictionary size $1024$ (expansion
factor $2$). Our obtained results suggest that overcomplete dictionaries are not
essential for the proposed anchoring mechanism, however, they can still be used
when interpretability of continual learning models representations is an
additional objective, while the same SAE can be used to derive the
Jacobian-based anchors for continual learning.

\begin{table}[h]
\centering
\caption{Effect of SAE dictionary size on accuracy for CIFAR-100 with a ResNet-18 trained from scratch, at fixed $K = r = 16$. Expansion factor is relative to the backbone's $512$-dimensional feature space. Results are mean $\pm$ std over three seeds. Higher results are better ($\uparrow$).}
\label{tab:latentdim_ablation}
\begin{tabular}{lccc}
\toprule
\textbf{Dictionary size} & \textbf{Expansion} & \textbf{Class-IL} & \textbf{Task-IL} \\
\midrule
256 
& $0.5\times$
& $\mathbf{28.53 \pm 0.52}$ & $\mathbf{74.35 \pm 0.43}$ \\
512
& $1\times$
& $26.67 \pm 0.87$ & $71.29 \pm 0.63$ \\
1024
& $2\times$
& $24.66 \pm 5.33$ & $69.49 \pm 4.39$ \\
2048
& $4\times$
& $28.07 \pm 1.41$ & $71.80 \pm 1.09$ \\
4096
& $8\times$
& $28.10 \pm 2.15$ & $71.51 \pm 1.22$ \\
\bottomrule
\end{tabular}
\end{table}

\section{Detailed experimental setup}\label{app:experimental_details}
\subsection{Datasets}\label{appsec:datasets_from_scratch}

Table~\ref{tab:datasets} summarizes the datasets. Classes are assigned to tasks in the class order of the dataset (no class permutation), so the task sequence is identical for every seed; each task is trained for a fixed number of epochs (Table~\ref{tab:alt_hyperparams}). CIFAR-100 is used at native $32\times32$ resolution with a from scratch ResNet-18 and upsampled to $224\times224$ with the pretrained ViT-B/16. TinyImageNet has $500$ training images per class and its official validation split serves as the test set. CUB-200 (fine-grained bird species) has about $30$ training images per class and ImageNet-R (renditions such as art and cartoons of $200$ ImageNet classes) $120$ per class, with the fixed train/test split of the respective data release. In all settings the seed changes the initialization of the new layers and of the SAE, the data order and the buffer contents, but not the task sequence.

\begin{table}[h]
\centering
\caption{Datasets. ``Tasks'' is the number of tasks $\times$ classes per task. Input: image size seen by the network. Train-time augmentation is listed in the text; test images are only resized (and, where stated, center-cropped) and normalized.}
\label{tab:datasets}
\resizebox{\linewidth}{!}{
\begin{tabular}{lcccccl}
\toprule
\textbf{Dataset} & \textbf{Classes} & \textbf{Tasks} & \textbf{Train / Test images} & \textbf{Input} & \textbf{Backbone} \\
\midrule
CIFAR-100 & 100 & $10\times10$ & 50,000 / 10,000 & $32\times32$ & ResNet-18 (scratch)  \\
TinyImageNet & 200 & $10\times20$ & 100,000 / 10,000 & $64\times64$ & ResNet-18 (scratch) \\
CIFAR-100 & 100 & $10\times10$ & 50,000 / 10,000 & $224\times224$ & ViT-B/16 (pretrained) \\
ImageNet-R & 200 & $10\times20$ & 24,000 / 6,000 & $224\times224$ & ViT-B/16 (pretrained) \\
CUB-200 & 200 & $10\times20$ & 5,994 / 5,794 & $224\times224$ & ViT-B/16 (pretrained)\\
\bottomrule
\end{tabular}}
\end{table}

\textbf{Datasets in training from scratch}
 A network trained from scratch needs many images per class. CIFAR-100 and TinyImageNet provide 500 training images per class, whereas ImageNet-R provides 120 (24,000 images for 200 classes, moreover renditions of varying style) and CUB-200 only 30 (5,994 images). In preliminary runs of a ResNet-18 trained from scratch on these two datasets (10 epochs per task, three seeds), all methods, including   plain SGD, stayed below 10\% Class-IL accuracy, so they cannot be meaningfully compared. In contrast, a ResNet-18 trained from scratch on TinyImageNet reaches 25--32\% with the rehearsal-based methods (and for SGD: 6.8\%). We therefore evaluate the from-scratch regime on CIFAR-100 and TinyImageNet and the pretrained regime on the 224-pixel datasets (upscaled CIFAR-100, ImageNet-R and CUB-200).

\subsection{Hyperparameters}\label{appsec:alt_hyperparams}

Table~\ref{tab:alt_hyperparams} lists the tuned hyperparameters of our method (learning rate and anchor weight) together with the fixed training settings, for the main results. The SAE-specific hyperparameters are fixed in all of them ($\texttt{sae\_epochs}=20$; $\texttt{sae\_lr}=10^{-3}$; dictionary size $256$).

\begin{table}[h]
\centering
\caption{Tuned hyperparameters and fixed training settings for our method SAGA (Jacobian + Replay + History), by setting.}
\label{tab:alt_hyperparams}
\resizebox{\linewidth}{!}{
\begin{tabular}{lccccccc}
\toprule
\textbf{Setting} & \textbf{Backbone} & \textbf{lr} & \textbf{Anchor weight} & \textbf{Epochs/task} & \textbf{Buffer} & \textbf{Optimizer} & \textbf{Batch} \\
\midrule
CIFAR-100 & ViT-B/16 (pretrained) & $3\times10^{-5}$ & $3.0$ & $10$ & $500$ & Adam & $64$ \\
ImageNet-R & ViT-B/16 (pretrained) & $3\times10^{-5}$ & $0.3$ & $10$ & $500$ & Adam & $64$ \\
CUB-200 & ViT-B/16 (pretrained) & $3\times10^{-5}$ & $0.3$ & $10$ & $500$ & Adam & $64$ \\
CIFAR-100 & ResNet-18 (scratch) & $0.1$ & $0.3$ & $20$ & $500$ & SGD & $64$ \\
TinyImageNet & ResNet-18 (scratch) & $0.03$ & $1.0$ & $20$ & $5120$ & SGD & $64$ \\
\bottomrule
\end{tabular}}
\end{table}

For every setting, our method's learning rate and anchor weight were selected by
a grid search (learning rate $\times$ anchor weight) evaluated on a held-out
validation split ($10\%$ of the training set held out per class), using a single
screening seed. All numbers reported elsewhere in the paper are then
produced by a separate run at the winning
hyperparameters, over three seeds on the separate test set.


\subsection{Comparisons with reference geometries}\label{appsec:ref_geomteruies}

To test whether SAE-Jacobian anchoring benefits from SAE-induced geometry specifically or from low-rank anchoring more generally, we compare it with random, PCA, and classifier SVD geometries under the same Replay+History protocol. Each geometry supplies an orthonormal basis $V_r$ and nonnegative direction weights $S_r$, defining
\begin{equation}
    Q_r^{(m)}=V_r^{(m)}S_r^{(m)2}V_r^{(m)\top}.
\end{equation}
For these comparisons we use \texttt{anchor\_weight\_norm=normalized}, which rescales the weights according to
\begin{equation}
\label{eq:singular_weight_normalization}
    \widetilde S_r
    =\frac{S_r}{\sqrt{\frac{1}{r}\sum_{i=1}^r S_i^2}}.
\end{equation}
Because the implementation also averages squared projected drift over the $r$ retained directions, the normalized per-example loss is
\begin{equation}
\label{eq:trace_normalized_geometry}
    \frac{1}{r}\|\widetilde S_rV_r^\top\delta\|_2^2
    =\frac{\delta^\top Q_r^{(m)}\delta}{\operatorname{tr}Q_r^{(m)}}.
\end{equation}
Thus the comparison normalizes the total metric trace while preserving each geometry's directions and relative within-subspace spectral weighting.

The concrete reference geometries are:
\begin{itemize}
    \item \emph{Random}: draw $r$ Gaussian directions and orthonormalize them by
    QR. The directions use unit weights, so $Q_r=V_rV_r^\top$ and the
    implemented loss is $\|V_r^\top\delta\|_2^2/r$. The seed of the basis is derived from the run seed and the task index.
    \item \emph{PCA}: center the same representation dataset that would
    otherwise be used to fit the SAE, compute $H_c=USV^\top$, and use the
    leading right-singular directions together with their singular values.
    Before normalization, $Q_r^{\mathrm{PCA}}=V_rS_r^2V_r^\top$.
    \item \emph{Classifier SVD}: decompose the frozen classifier weight matrix
    $W=USV^\top$ and use its leading right-singular directions and singular
    values. Before normalization, the full-rank energy satisfies
    $\|SV^\top\delta\|_2^2=\|W\delta\|_2^2$, i.e. squared drift of a frozen
    linear readout. Under Eq.~\ref{eq:trace_normalized_geometry}, the
    implemented full-rank quantity is proportional to logit drift,
    $\|W\delta\|_2^2/\|W\|_F^2$. The requested rank is capped by the numerical
    rank of $W$.
    \item \emph{SAE Jacobian}: use the per-example right-singular directions and
    singular values of the local SAE encoder Jacobian. Unlike the other three
    controls, this basis varies across anchor examples.
\end{itemize}
PCA and classifier SVD therefore vary both the protected directions and their
within-subspace spectral weights. The experiment should be interpreted as a
comparison of alternative low-rank geometries rather than of directions alone.

\subsection{SAE setup}\label{appsec:sae_setup}

At every task boundary a fresh TopK SAE \cite{gao2025scaling} is trained on the frozen backbone's features of the reservoir snapshot; its encoder Jacobian is evaluated at the snapshot's samples and, in the Jacobian mode used for all main results, the SAE is then discarded, keeping only the low-rank directions and weights.

\textbf{Architecture.} SAEs act on the pooled pre-logit feature $h\in\mathbb{R}^D$ (class token after the final layer normalization for ViT-B/16, $D=768$; globally average-pooled last-stage output for ResNet-18, $D=512$), used without centering or rescaling. It has one hidden layer with a hard TopK activation,
\begin{equation}
\label{eq:sae_architecture}
    E(h)=\operatorname{TopK}_K\!\big(\operatorname{ReLU}(W_eh+b_e)\big),
    \qquad
    \hat h=W_dE(h)+b_d,
\end{equation}
with $W_e\in\mathbb{R}^{m\times D}$ and $W_d\in\mathbb{R}^{D\times m}$; $\operatorname{TopK}_K$ keeps the $K$ largest post-ReLU activations of each sample in place and zeroes the rest. The weights are untied, the decoder columns are unconstrained, initialization is the PyTorch default, and there are $2Dm+m+D$ parameters. The values used for the main results are listed in Table~\ref{tab:sae_setup_configs}: $m=256$ and $K=16$ (undercomplete: $0.33\times$ the feature dimension for ViT-B/16, $0.5\times$ for ResNet-18). The ablations change them as follows: $K=64$ for all ranks $r\in\{4,8,16,32,64\}$ in the rank ablation, so that the SAE is identical across ranks; $K=r$ in the reference-geometry comparison, except $K=16$ at $r=4$; and $m\in\{256,\ldots,4096\}$ at $K=r=16$ in the dictionary-size ablation.

\textbf{Training.} The SAE minimizes only the mean squared reconstruction error over minibatches $\mathcal{B}$,
\begin{equation}
\label{eq:sae_loss}
    \mathcal{L}_{\mathrm{SAE}}
    =\frac{1}{|\mathcal{B}|\,D}\sum_{h\in\mathcal{B}}\big\|\hat h-h\big\|_2^2,
\end{equation}
with no sparsity penalty (TopK provides the sparsity), auxiliary loss or weight decay. It uses Adam (learning rate $10^{-3}$, default momenta) for $20$ epochs with batch size $64$, reshuffled every epoch, independently of the continual learner's optimizer. It is initialized from scratch at every boundary and then frozen in evaluation mode: $8$ steps per epoch ($160$ in total) for $500$ samples, $1600$ steps for the $5120$ samples of TinyImageNet. These values are identical in all settings and were not tuned.

\textbf{Training data.} The SAE is trained without labels on the features of all $N=\min(\text{images seen},\text{buffer capacity})$ images in the reservoir buffer at the boundary ($N\le500$; $N\le5120$ for TinyImageNet). As in the baselines, the reservoir is updated at every training step, so it approximates a uniform sample of the stream seen so far, contains all classes seen up to that task, and, when a task is trained for several epochs, may hold the same image more than once. Images are stored without augmentation and passed through the deterministic part of the dataset's test-time preprocessing (resizing, normalization), as when the anchor loss compares $h_{\mathrm{new}}$ with $h_{\mathrm{old}}$; features come from the backbone at the boundary in evaluation mode. 

\textbf{From the SAE to the geometry.} For a buffer image $x$ with feature $h_{\mathrm{old}}$, let $\pi_1(x),\ldots,\pi_K(x)$ index its $K$ largest post-ReLU activations $a=\operatorname{ReLU}(W_eh_{\mathrm{old}}+b_e)$. Away from ReLU and TopK boundaries the encoder is locally linear and its Jacobian is read from the weights in closed form:
\begin{equation}
\label{eq:sae_jacobian_rows}
    J_x=\big[\,\mathbb{1}[a_{\pi_i(x)}>0]\;(W_e)_{\pi_i(x),:}\,\big]_{i=1}^{K}\in\mathbb{R}^{K\times D}.
\end{equation}
We take the thin SVD $J_x=U\Sigma V^\top$ (in chunks of $64$ samples), keep the $r\le K$ leading $\sigma_i$ and $v_i$, and store them with $h_{\mathrm{old}}$, the image and its label in the snapshot; the normalized weighting rescales $\sigma$ to $\tilde\sigma$ (Eq.~\ref{eq:singular_weight_normalization}). In Latent mode the frozen SAE is kept and $E(h_{\mathrm{old}})\in\mathbb{R}^m$ is stored instead. Anchors therefore need only the buffer images, the backbone at the boundary and the SAE (transiently, while the directions are computed); no data of the finished task, old-task classifier or gradients are used. The reference geometries of Section~\ref{sec:sanity_checks} train no SAE: \emph{random} needs only $D$, $r$ and a seed (Gaussian matrix orthonormalized by QR; seed derived from the run seed and task index), \emph{PCA} the centered buffer features (leading right singular vectors and values), \emph{classifier SVD} the classifier weights at the boundary. These bases are global, one $D\times r$ matrix shared by all anchors, whereas SAE-Jacobian directions are per anchor.

\begin{table}[h]
\color{black}
\centering
\caption{SAE hyperparameters for the main results. $D$: feature dimension; $m$: dictionary size; $K$: TopK sparsity; $r$: retained Jacobian rank; $N$: buffer samples per snapshot (SAE training set and anchors); SAE parameters: $2Dm+m+D$. The training hyperparameters (Adam, learning rate $10^{-3}$, $20$ epochs, batch size $64$) are identical in all settings.}
\label{tab:sae_setup_configs}
\scriptsize
\begin{tabular}{lcccccc}
\toprule
\textbf{Backbone; datasets} & $D$ & $m$ & $K$ & $r$ & $N$ & \textbf{SAE params} \\
\midrule
ViT-B/16; CIFAR-100, CUB-200, ImageNet-R & 768 & 256 & 16 & 16 & 500 & 394k \\
ResNet-18; CIFAR-100 & 512 & 256 & 16 & 16 & 500 & 263k \\
ResNet-18; TinyImageNet & 512 & 256 & 16 & 16 & 5120 & 263k \\
\bottomrule
\end{tabular}
\end{table}

\FloatBarrier

\section{Computational Overhead}\label{sec:alt_timing}

Table~\ref{tab:alt_timing} reports the mean wall-clock training time per task
for our method and SGD, included as the least computationally expensive
baseline, across five different settings for which per-task timing was
recorded. Independently measured GPU-device time using CUDA events differs from
wall-clock time by less than $1\%$ in all cases, indicating that the runs are
predominantly GPU-bound with negligible CPU and data-loading overhead. We
therefore omit GPU-device time as a separate column. 

\begin{table}[h]
\centering
\caption{Mean wall-clock time per task (seconds), our method vs.\ SGD, across the {\color{black}five} settings with timing instrumentation. Ratio = Ours / SGD. }
\label{tab:alt_timing}
\begin{tabular}{lccc}
\toprule
\textbf{Setting} & \textbf{Ours (s/task)} & \textbf{SGD (s/task)} & \textbf{Ratio} \\
\midrule
CUB-200 (ViT-B/16, pretrained) & $114.3$ & $16.6$ & $6.9\times$ \\
ImageNet-R (ViT-B/16, pretrained) & $429.8$ & $67.0$ & $6.4\times$ \\
{\color{black}CIFAR-100 (ViT-B/16, pretrained)} & {\color{black}$904.2$} & {\color{black}$138.6$} & {\color{black}$6.5\times$} \\
{\color{black}CIFAR-100 (ResNet-18, scratch)} & {\color{black}$82.6$} & {\color{black}$12.9$} & {\color{black}$6.4\times$} \\
{\color{black}TinyImageNet (ResNet-18, scratch)} & {\color{black}$302.2$} & {\color{black}$42.8$} & {\color{black}$7.1\times$} \\
\bottomrule
\end{tabular}
\end{table}

The proposed method introduces a consistent computational overhead across all
evaluated settings, with training time remaining within a narrow
$6.4\times$--$7.1 \times$ range relative to SGD. This indicates that the
additional cost is predictable and primarily comes from SAE training,
Jacobian/SVD computation at task boundaries, and the additional anchoring passes
during optimization.

\end{document}